\documentclass{article}
\usepackage{spconf,amsmath,graphicx}

\title{Embedding Semantic Hierarchy in Discrete Optimal Transport\\ for Risk Minimization}
\usepackage{times}
\usepackage{epsfig}
\usepackage{graphicx}
\usepackage{amsmath}
\usepackage{amssymb}
\usepackage{algorithm}
\usepackage{algorithmic}
\usepackage{mathrsfs}
\usepackage{multirow}

\usepackage{amsthm}
\theoremstyle{plain}

\usepackage[section]{placeins}

\usepackage[]{animate}

\usepackage{graphicx}
\usepackage{animate}

\usepackage{float}

\usepackage{multirow}
\usepackage{rotating}
\usepackage{wrapfig}
\usepackage{lineno}
\usepackage{epsfig}
\usepackage{amsmath,amssymb}
\usepackage{color}
\usepackage{booktabs}       
\usepackage{amsfonts,amsmath}       
\usepackage{nicefrac}       
\usepackage{microtype} 
\usepackage{float}
\usepackage{amsmath}
 
\usepackage{enumerate}

\name{Yubin Ge$^{1,2\dag}$, Site Li$^{1,3\dag}$, Xuyang Li$^{1,4}$, Fangfang Fan$^{1}$, Wanqing Xie$^{1}$, Jane You$^{3}$, Xiaofeng Liu$^{1,6*}$}
\address{$^{1}$Harvard University $^{2}$University of Illinois at Urbana-Champaign $^{3}$Carnegie Mellon University \\$^{4}$Northeastern University, $^{5}$Hong Kong Polytechnic University, $^{6}$Fanhan Tech\\$^\dag$Contribute equally. $^*$Corresponding author: liuxiaofengcmu@gmail.com} 

\begin{document}
%
\maketitle
\begin{abstract}
The widely-used cross-entropy (CE) loss-based deep networks achieved significant progress w.r.t. the classification accuracy. However, the CE loss can essentially ignore the risk of misclassification which is usually measured by the distance between the prediction and label in a semantic hierarchical tree. In this paper, we propose to incorporate the risk-aware inter-class correlation in a discrete optimal transport (DOT) training framework by configuring its ground distance matrix. The ground distance matrix can be pre-defined following a priori of hierarchical semantic risk. Specifically, we define the tree induced error (TIE) on a hierarchical semantic tree and extend it to its increasing function from the optimization perspective. The semantic similarity in each level of a tree is integrated with the information gain. We achieve promising results on several large scale image classification tasks with a semantic tree structure in a plug and play manner.
\end{abstract}
\section{Introduction}

Conventionally, the risk minimization in deep learning is based on $N$-way flat softmax prediction and cross-entropy (CE) loss, where $N$ is the number of categories. However, it can ignore the correlation of different classes and can not discriminate different kinds of misclassification \cite{liu2018ordinal,liu2020unimodal}.

\begin{figure}[t]
\centering
\includegraphics[height=5.5cm]{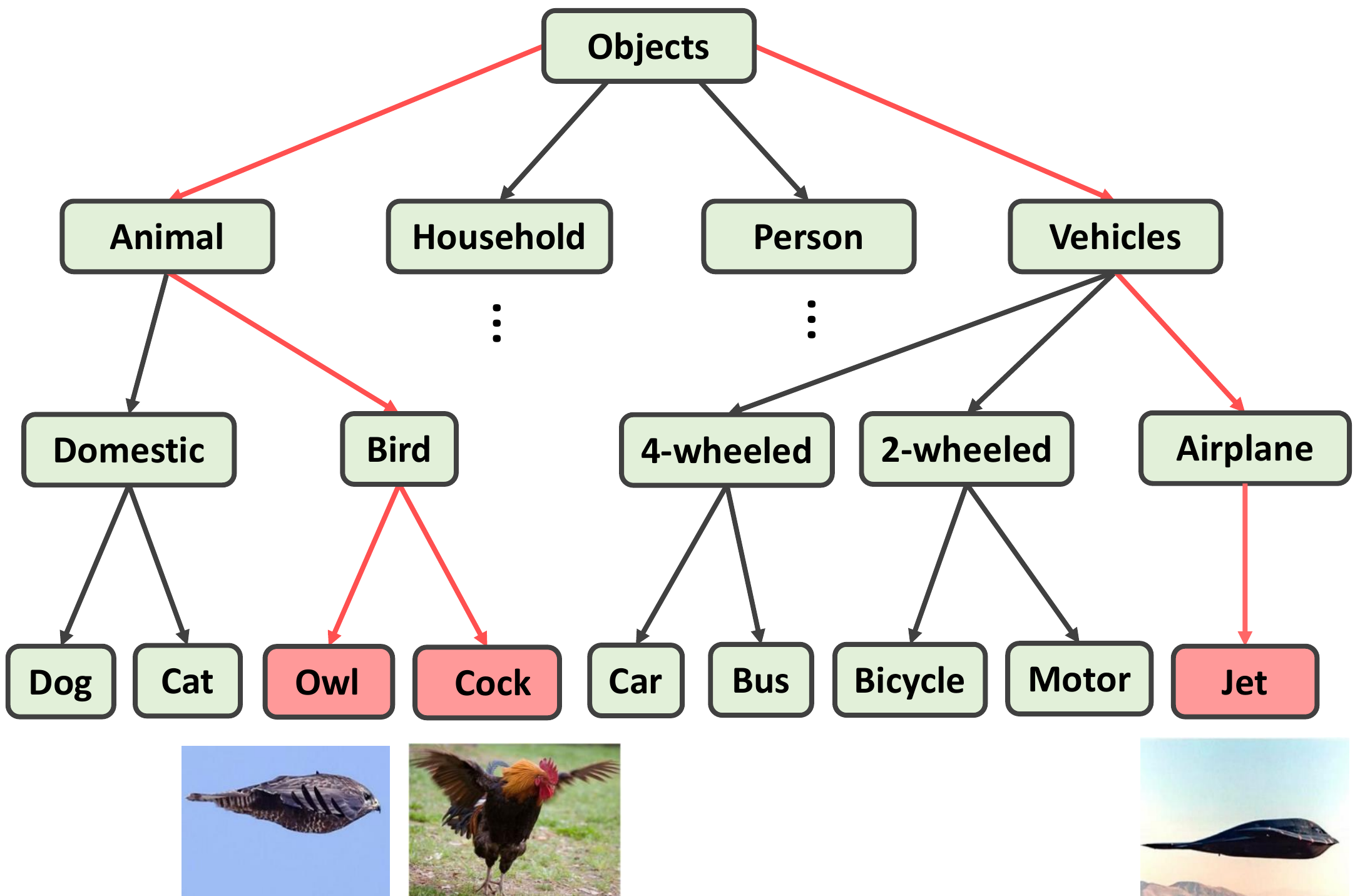} 
\caption{Simplified illustration of the semantic hierarchy in ImageNet. Misclassifying a cock as an owl or a jet are both counted as one mistake with CE loss, where the degree of mistakes is overlooked. The semantic hierarchy provides a useful cue to differentiate such risk difference.} 
\label{fig:1}
\end{figure} 

Actually, there is an inherent semantic tree structure in the label manifold for many large scale datasets \cite{ceci2007classifying}. As shown in Fig. \ref{fig:1}, the cock and owl should have the smaller semantic distance than the cock with jet, and the semantic correlation can be explicitly defined by the semantic hierarchical tree structure. With the prior of concept ontology, we are able to incorporate the degree of risks of mistakes, which is of great significance in many real-world tasks \cite{wang2020hierarchical}. Moreover, it can be used for making coarse-to-fine prediction to rule out unlikely groups of classes and therefore also benefit for the overall recognition accuracy \cite{zhao2018embedding}.

The progress for semantic hierarchy-aware risk minimization made by recent works either configure a sophisticate network structure or design a complicate inference logic that suffers from the effective optimization algorithms \cite{wang2017local,lee2018hierarchical,khan2017cost}.

\begin{figure}[t]
\centering
\includegraphics[height=3.2cm]{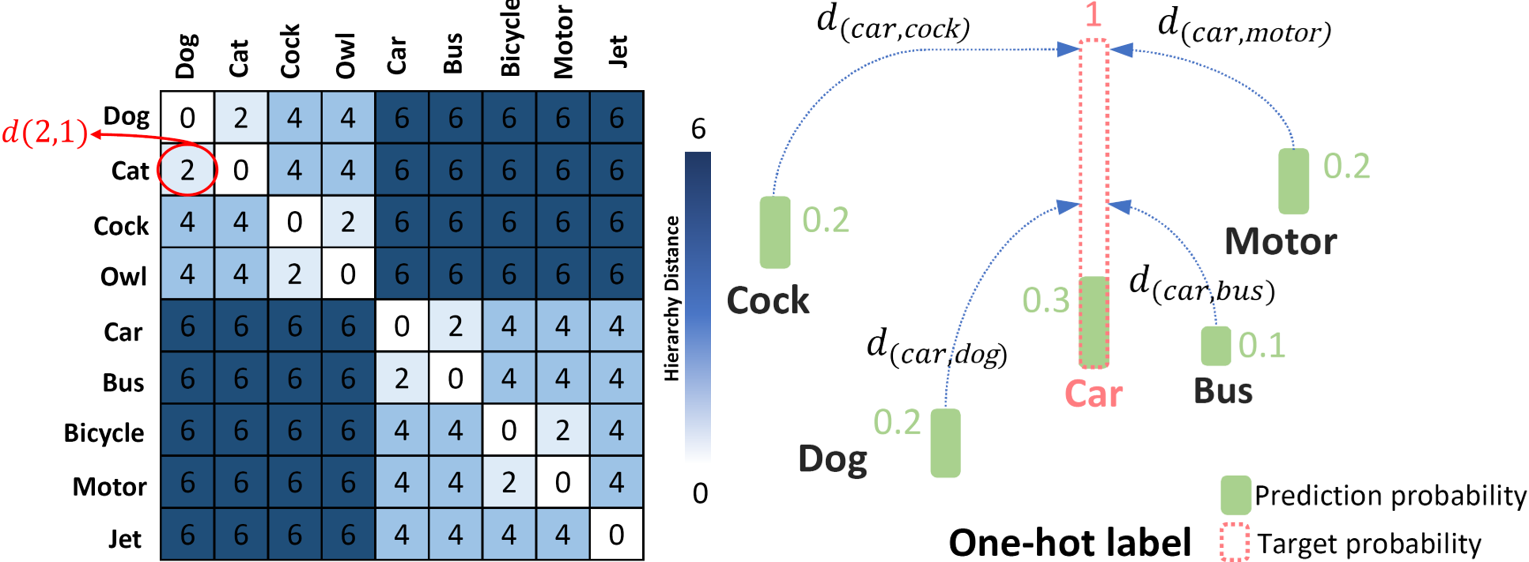} 
\caption{(a) The ground matrix of a hierarchical tree with TIE. (b) The only possible transport plan in one-hot target case.} 
\label{fig:3}
\end{figure}

In this paper, we resort to the optimal transport distance as an alternative for empirical risk minimization \cite{liu2019unimodala,liu2019conservative,liu2020importance,liu2020wasserstein}. With the low-cost modification of the loss function perspective, our solution can be added on any up-to-date general deep networks in a plug-and-play fashion. The distance is defined as the cost of optimal transport for moving the mass in one distribution to match the target distribution \cite{liu2020severity,han2020wasserstein}. Specifically, we measure the discrete optimal transport distance between a softmax prediction and its target label, both of which are normalized as histograms. By defining the ground metric as semantic similarity, we can measure prediction in a way that is sensitive to the semantic correlations between the classes in a tree. We design a ground matrix utilizing the semantic tree structure. The ground metric can be predefined when the similarity structure is known a priori to incorporate the inter-class correlation, $e.g.,$ the tree induced error (TIE) \cite{silla2011survey}. Formally, TIE$=|L_y|+|L_{\hat{y}}|-2|L_y\cap L_{\hat{y}}|$, where $|L_y|$ and $|L_{\hat{y}}|$ are the number of link between root node to the ground truth node and predict node respectively, $|L_y\cap L_{\hat{y}}|$ is the number of common link \cite{kosmopoulos2015evaluation}. The TIE in Fig \ref{fig:1} is shown in Fig \ref{fig:3} left.

We configure a multi-task network to explore the semantic similarity at each level. Since the misclassification in the high-level could be more severe than the low-level, the learning objective of the shared trunk network is balanced by the information gain \cite{deng2012hedging} of each level.

In summary, we cast the semantic hierarchy recognition as a discrete optimal transportation training problem. The inter-class relationship of each class is explicitly incorporated as prior information in our ground metric which can be pre-defined as a function $w.r.t.$ hierarchy tree. We empirically validate the effectiveness and generality of the proposed method on multiple challenging benchmarks and achieve state-of-the-art performance.

\section{Methodology}

We target to learn a classifier ${h}_\theta$, parameterized by $\theta$, with a conventional softmax output unit. It projects an inquire image ${\rm\textbf{X}}\in\mathbb{R}^{h\times w\times 3}$ to a vector ${\rm\textbf{s}}\in\mathbb{R}^N$, where $N$ is the number of pre-defined classes in a level of hierarchy, $h$ and $w$ are the height and width of input image respectively. 

Let ${\rm\textbf{s}}=\left\{s_i\right\}_{i=1}^{N}$ be the prediction of a image in ${h}_\theta({\rm\textbf{X}})$, $i.e.,$ softmax normalized $N$ classes probability. $i\in\left\{1,\cdots,{\small N}\right\}$ be the index of dimension (class). We perform learning over a hypothesis space $\mathcal{H}$ of ${h}_\theta$. 

Given input {\rm\textbf{x}} and its target ground truth one-hot label ${\rm\textbf{t}\in\mathbb{R}^{N}}$, typically, learning is performed via empirical risk minimization to solve $\mathop{}_{{h}_\theta\in\mathcal{H}}^{\rm min}\mathcal{L}({h}_\theta({\rm\textbf{x}}),{\rm\textbf{t}})$, with a loss $\mathcal{L}(\cdot,\cdot)$ acting as a surrogate of performance measure.

Unfortunately, cross-entropy, information divergence, Hellinger distance and $\mathcal{X}^2$ distance-based loss treat the output dimensions independently \cite{frogner2015learning}, ignoring the similarity structure on label space.

Optimal Transport Distance is the cost of moving the mass from source to target distribution, which is related to the weight of mass and the moving distance \cite{villani2003topics}. Let define ${\rm\textbf{t}}=\left\{t_j\right\}_{j=1}^{N}$ as the target histogram distribution label that can be either one-hot or non-one-hot vector. Assume the class label possesses a ground metric ${\rm\textbf{D}}_{i,j}$, which measures the severity of misclassifying $i$-th class pixel into $j$-th class. There are $N^2$ possible ${\rm\textbf{D}}_{i,j}$ in a $N$ class dataset and form a ground distance matrix $\textbf{D}\in\mathbb{R}^{N\times N}$ \cite{ruschendorf1985wasserstein}. When ${\rm\textbf{s}}$ and ${\rm\textbf{t}}$ are both histograms, the discrete measure of exact optimal transport loss is defined as \begin{equation}
\mathcal{L}_{\textbf{D}_{i,j}}({\rm{\textbf{s},\textbf{t}}})=\mathop{}_{\textbf{T}}^{{\rm inf}}\sum_{j=0}^{N-1}\sum_{i=0}^{N-1}\textbf{D}_{i,j}\textbf{T}_{i,j} \label{con:df}
\end{equation} where \textbf{T} is the transportation matrix with \textbf{T}$_{i,j}$ indicating the mass moved from the $i^{th}$ point in source distribution to the $j^{th}$ target position. $\mathbf{D}_{i,j}$ indicates the element in ground matrix $\mathbf{D}$, its value equals to the ground distance $f(d_{i,j})$, where $d_{i,j}$ is TIE distance in this paper. 

A valid transportation matrix \textbf{T} satisfies: $ \textbf{T}_{i,j}\geq 0$; $\sum_{j=0}^{N-1} \textbf{T}_{i,j}\leq s_i$; $\sum_{i=0}^{N-1} \textbf{T}_{i,j}\leq t_j$; $\sum_{j=0}^{N-1}\sum_{i=0}^{N-1} \textbf{T}_{i,j}={\rm min}(\sum_{i=0}^{N-1}s_i,\sum_{j=0}^{N-1}t_j)$. 

A possible ground distance matrix ${\rm\textbf{D}}$ in our application is the tree induced error(TIE) as shown in Fig. \ref{fig:3}. TIE equals the number of edges in the link from one node to another. For instance, classifying the car to the cat ($d_{2,5}$) has a larger ground distance than car to bus ($d_{2,4}$). Here we use the symmetric distance $d_{i,j}$ as ${\rm\textbf{D}}_{i,j}$. Although the entries in matrix ${\rm\textbf{D}}$ is not necessary to be symmetric with respect to the main diagonal in our setting that the matrix is adaptively learned.

\subsection{Optimal transport with one-hot target}

The one-hot encoding is a typical setting for multi-class one-label dataset. The distribution of a target label probability is ${\rm\textbf{t}}=\delta_{j,j^*}$, where $j^*$ is the ground truth class, $\delta_{j,j^*}$ is a Dirac delta, which equals to 1 for $j=j^*$\footnote{\noindent We use $i,j$ interlaced for ${\rm \textbf{s}}$ and ${\rm \textbf{t}}$, since they index the same group of categories.}, and $0$ otherwise.

\vspace{+5pt}
\noindent\textbf{Theorem 1.} \textit{Assume that} $\sum_{j=0}^{N-1}t_j=\sum_{i=0}^{N-1}s_i$, \textit{and} ${\rm{\textbf{t}}}$ \textit{is a one-hot distribution with} $t_{j^*}=1 ($or $\sum_{i=0}^{N-1}s_i)$\footnote{We note that softmax cannot strictly guarantee the sum of its outputs to be 1 considering the rounding operation. However, the difference of setting $t_{j^*}$ to $1$ or $\sum_{i=0}^{N-1}s_i$ is not significant in our experiments using the typical format of softmax output which is accurate to 8 decimal places.}, \textit{there is only one feasible optimal transport plan.}  

\vspace{+5pt}
\noindent\textbf{Sketch proof of Algorithm 1.} Following \cite{cuturi2013sinkhorn}:

$\mathcal{L}_{\textbf{D}_{i,j}}({\rm{\textbf{s},\textbf{t}}})=\mathop{}_{\textbf{T}\in\Pi(\textbf{s},\textbf{t})}^{{\rm ~~~inf}}  \left \langle \textbf{D},\textbf{T}  \right \rangle $ and the set of valid transport plan is $\Pi(\textbf{s},\textbf{t})=\{\textbf{T}\in^{N\times N}_{+}: \textbf{T}\textbf{1}=\textbf{s},\textbf{T}^{\top}\textbf{1}=\textbf{t}$\}, where $\textbf{1}$ is the all-one vector, $\textbf{T}$ is the transportation matrix. \vspace{+5pt}

Given the label $\textbf{t}$ is a ``one-hot'' vector $\textbf{t}_j^*$, with only $t_j^*=1$ (0 for the others), the constraint $\textbf{T}^{\top}\textbf{1}=\textbf{t}_j^*$ means that only the $j^*$-th column of $\textbf{T}$ can be non-zero. Furthermore, the constraint $\textbf{T}\textbf{1}=\textbf{s}$ ensures that the $j^*$-th column of $\textbf{T}$ actually equals $h_{{\theta}}(\cdot|x)=\textbf{s}$. In other words, the set $\Pi(\textbf{s},\textbf{t})$ contains only one feasible transport plan, that is to move all of the source probability to the $j^*$-th place. Therefore, Eq. 1 can be computed directly as $\sum_{i=0}^{N-1} s_i \textbf{D}_{i,j^*} =\sum_{i=0}^{N-1} s_i f(d_{i,j^*})$.

Actually, it is easy to understand intuitively. Suppose the target distribution is \{0,0,1,0\} (i.e., $j^*=3$), and the source is \{0.2,0.2,0.5,0.1\}, the only optimal route is simply removing 0.2,0.2,0.1 to the third class.    

According to the criteria of ${\rm \textbf{T}}$, all masses have to be transferred to the cluster of the ground truth label $j^*$, as illustrated in Fig. \ref{fig:3}. Then, the optimal transport distance between softmax prediction {\rm{\textbf{s}}} and one-hot target {\rm{\textbf{t}}} degenerates to\begin{equation}
\mathcal{L}_{{\rm\textbf{D}}_{i,j}^{f}}({\rm{\textbf{s},\textbf{t}}})=\sum_{i=0}^{N-1} s_i f(d_{i,j^*}) \label{con:df}
\end{equation} We propose to extend the ground metric in ${\rm\textbf{D}}_{i,j}$ as $f(d_{i,j})$, where $f$ can be a linear or increasing function proper, $e.g., p^{th}$ power of $d_{i,j}$ and Huber function from the optimization perspective. The exact solution of Eq. \eqref{con:df} can be computed with a complexity of $\mathcal{O}(N)$. The ground metric term $f(d_{i,j^*})$ works as the weights $w.r.t.$ $s_i$, which takes all classes into account following a soft attention scheme \cite{liu2018dependency}. It explicitly encourages the probabilities distributing on the neighboring classes of $j^*$. Since each $s_i$ is a function of the network parameters, differentiating $\mathcal{L}_{{\rm\textbf{D}}_{i,j}^{f}} w.r.t.$ network parameters yields $\sum_{i=0}^{N-1}s_i'f(d_{i,j^*})$.

In contrast, the CE loss in one-hot setting can be formulated as $-1{\rm log}s_{j^*}$. Similar to the hard prediction scheme, only a single class prediction is considered resulting in a large information loss \cite{liu2018dependency}. Besides, the regression loss with softmax prediction could be $f(d_{i^*,j^*})$, where $i^*$ is the class with maximum prediction probability.

\begin{figure}[t!]
\includegraphics[width=9cm]{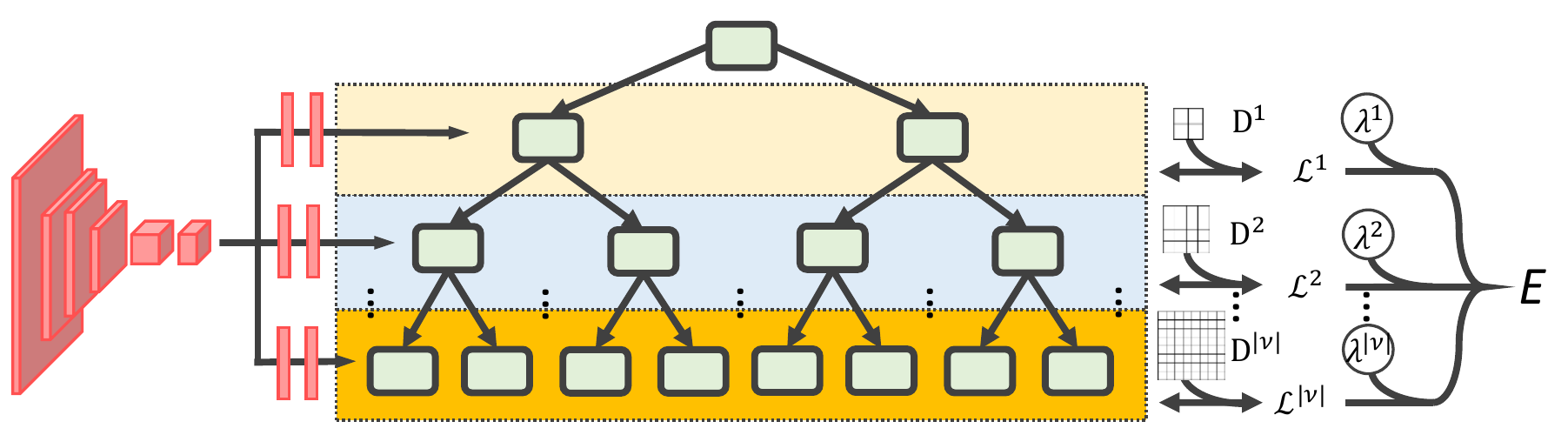} 
\caption{Illustration of the multi-task DOT framework.} 
\label{fig:44}
\end{figure}

\subsection{Implementation details}

The conventional hierarchical process train a large number of local classifier with the extracted features on each non-leaf node of the tree \cite{zhao2018embedding}. In this work, we leverage the idea of multi-task learning to simultaneously train classifiers for different levels. Inspired by the network architecture in \cite{ma2018end-to-end}, we share the convolutional features for all the $L$ tasks to enhance the ability of more discriminative feature learning for CNN model. In each task, we use the same fully connected layers structure for simply. In optimization, the loss function in each level is set as our optimal transport loss. The framework is illustrated in Figure \ref{fig:44}.

The combined loss for the shared convolutional network is defined as $E=\sum_l^L\lambda^l\mathcal{L}^l$, where $\lambda^l$ is a weight for each hierarchy level. It can be simply set to 1 and combine each level equally. However, the misclassification in the high-level could be more severe than the low-level. We develop from the concept of information gain \cite{deng2012hedging} as the decrease in number of leaf nodes to measure the importance of each level. Specifically, \begin{equation}
\lambda^l={\rm log}\{{|\mathcal{V}^l|}\}-{\rm log}\{{|\mathcal{V}|}\}
\end{equation} where ${|\mathcal{V}|}$ indicates the number of level in the pre-defined tree, and $|\mathcal{V}^l|={|\mathcal{V}|}-l$. At the inference stage, we simply use the fully connected branch of the leaf level, and give the output with argmax function. We note that the downpour algorithm \cite{wu2017hinet} may also be applied to find the maximum a posteriori (MAP) trace in the hierarchical tree, which models the highest probability of a trace among all possible route in the tree.

\begin{table}[]
\centering\caption{Experimental results of TIE metric (lower is better).} \vspace{+5pt}
\resizebox{1\linewidth}{!}{%
\begin{tabular}{c|c|c|c|c}
\hline
Method & PASCAL  VOC   & Stanford Cars & Caltech256 & ImageNet 1k \\ \hline\hline
TSS&  1.975 & 2.895  &  1.612 & 3.287 \\
DARTS&  1.898 &  2.838 & 1.598  & 2.937  \\
TKDL& 1.884  & 2.763  &  1.545 &  2.846\\
CSMSE& 1.826  &  2.884 &  1.605 & 3.051 \\
RMGA& 1.891  & 2.802  &  1.539 & 2.825 \\
HSRM &  1.840 & 2.732  & 1.498  &  2.798 \\ \hline

$\mathcal{L}_{d_{i,j}}$&1.722&2.637&1.365&2.710\\

$\mathcal{L}_{d_{i,j}}$leaf-only&1.751&2.653&1.392&2.735\\

$\mathcal{L}_{d_{i,j}}$equal&1.736&2.667&1.385&2.729\\\hline

\end{tabular}%
} 

\label{tab:1}
\end{table}

\begin{table}[t] 
\centering \caption{Comparison with no hierarchy method w.r.t. TIE($\downarrow$).}\vspace{+5pt}
\resizebox{1\linewidth}{!}{%
\begin{tabular}{c|c|c|c|c}
\hline
Method & PASCAL  VOC   & Stanford Cars & Caltech256 & ImageNet 1k \\ \hline\hline
Softmax+CE&  2.358$\pm$0.013 & 3.091$\pm$0.011  &  1.836$\pm$0.009 & 3.522$\pm$0.016 \\\hline

$\mathcal{L}_{d_{i,j}}$leaf-only&1.751$\pm$0.010&2.653$\pm$0.009&1.392$\pm$0.011&2.735$\pm$0.011\\

$\mathcal{L}_{d_{i,j}}$equal&1.736$\pm$0.011&2.667$\pm$0.011&1.385$\pm$0.008&2.729$\pm$0.012\\

$\mathcal{L}_{d_{i,j}}$&1.722$\pm$0.009&2.637$\pm$0.010&1.365$\pm$0.010&2.710$\pm$0.009\\\hline
\end{tabular}}
\label{tab:11}
\end{table}

\section{Experiments}

\noindent\textbf{Datasets.}

We evaluate our method on four challenging hierarchical classification benchmarks with semantic tree structure, including PASCAL VOC (34828 samples, 20 leafs, $|\mathcal{V}|$=5) \cite{everingham2010pascal}, Stanford Cars (16185 samples, 196 leafs, $|\mathcal{V}|$=3) \cite{krause2014learning}, Caltech256 (30607 samples, 256 leafs, $|\mathcal{V}|$=3) \cite{griffin2007caltech}, ImageNet 1k (1321167 samples, 1000 leafs, $|\mathcal{V}|$=19) and ImageNet 10k. For example, Stanford cars with the level of Make, Model, Year (Tesla$\rightarrow$Model S$\rightarrow$2012 model).

It is worth mentioning that these databases can test the models in various perspectives. For example,  the Stanford Cars database consists of different cars in fine-grained classes, which is challenging to differentiate. The ImageNet databases have massive labels, and has a very complicated semantic tree structure with 19 levels. We use all these databases to provide comprehensive testing. 

\vspace{+5pt}
\noindent\textbf{Evaluation Metrics.}

To evaluate the performance, rather than accuracy, we also use classic hierarchical evaluation metrics tree induced error (TIE) \cite{ceci2007classifying} to measure the performance for intuitive understanding and fair comparison.

\vspace{+5pt}
\noindent\textbf{Numerical results and ablation study.}

In all of the experiments, we use the Adam optimization and the mini-batch of 128. The learning rate is set to $\alpha=10^{-4}$ or $10^{-5}$ in different tasks according to previous work.

We follow the standard training and testing protocol \cite{wang2020hierarchical,wang2017local}, all the databases are split into training set, validation set and test set by 50\%, 30\%, 20\%, respectively. The training set is used to train multi-task hierarchical classifier and adaptively learn the ground metric, the validation set is applied to tune the hyper-parameters and the test set is applied to obtain the test results. All of the results shown are the average of 10 trails of random splitting. We implement our methods using the PyTorch toolbox.

\begin{table}[]
\centering\caption{Experimental results of Top-1 accuracy ($\uparrow$).}\vspace{+5pt}
\resizebox{1\linewidth}{!}{%
\begin{tabular}{c|cccccc|c}
\hline
Method & PASCAL  VOC   & Stanford Cars & Caltech256 & ImageNet 1k \\ \hline\hline

RMGA& 45.32\%  &  54.76\% & 83.74\%   & 57.38\% \\
HSRM & 45.35\% & 54.88\%  & 84.01\%&57.85\%  \\ \hline

$\mathcal{L}_{d_{i,j}}$& 46.25\%  &56.26\%&  85.32\%  & 58.83\% \\

$\mathcal{L}_{d_{i,j}}$leaf-only&45.98\%&56.15\%&84.93\%&58.25\%\\

$\mathcal{L}_{d_{i,j}}$equal&46.06\%&56.23\%&85.14\%&58.32\%\\\hline

\end{tabular}%
} 

\label{tab:2}
\end{table}

\begin{table}[] 
\centering\caption{The top-k error ($\downarrow$) comparison on ImageNet 10K with the same protocol and backbone network in LMM.}\vspace{+5pt}
\resizebox{1\linewidth}{!}{%
\begin{tabular}{c|c|c|c|c}
\hline
&\multicolumn{4}{c}{Top-k prediction error}\\\cline{2-5}

Method&1&2&5&10\\\hline\hline

Vanilla (softmax+CE) &70.30\%&60.33\%&47.99\%&39.20\%\\
LMM\cite{zhao2018embedding} &69.50\%&59.39\%&46.88\%&38.07\%\\

$\mathcal{L}_{d_{i,j}}$&68.22\%&58.37\%&45.65\%&37.10\%\\\hline

\end{tabular}%
} 
\label{tab:3}
\end{table}

We compare the proposed model with classic and state-of-the-art algorithms, including TSS \cite{ceci2007classifying}, DARTS \cite{deng2012hedging}, RMGA \cite{wang2017local}, TDKL \cite{lee2018hierarchical}, CSMSE \cite{khan2017cost} and HSRM \cite{wang2020hierarchical}. 

To comprehensively compare the performance of all the models, we show the results of TIE on four databases in Table \ref{tab:1}. For cost-sensitive method CSMSE \cite{khan2017cost}, we implement the end-to-end version by using the Mean Square Error (MSE) loss in the paper. We note that the TSS \cite{ceci2007classifying}, DARTS \cite{deng2012hedging} and RMGA \cite{wang2017local} are developed under the condition that a local classifier is trained on each non-leaf node, while TDKL \cite{lee2018hierarchical} and HSRM \cite{wang2020hierarchical} can jointly optimize the tree classifier.


As shown in Table \ref{tab:1}, our proposed discrete optimal transportation (DOT) loss can significantly outperform previous methods w.r.t. TIE. This indicates that misclassification made by DOT may less severe. For the very confusing case (unavoidable to make mistakes), the classifier is more likely to misclassify the object to the class with closer semantic distance.  Even DOT and conventional CE loss have the similar probability to be wrong (i.e., same argmax softmax probability), their consequences will have different severity. We also add the CE loss version in Tab. 2 (with the same backbone) for fair comparison.

We note that DOT does NOT explicitly targets for the higher accuracy, but manages to minimize TIE. Although the semantic classification may making coarse-to-fine prediction to rule out unlikely groups of classes and therefore also benefit for the overall recognition accuracy \cite{zhao2018embedding}. As shown in Table \ref{tab:2}, our method does not sacrifice the accuracy to achieve a better TIE metric. By introducing a more strict optimization objective, the proposed DOT can usually achieve higher accuracy or on par with the other STOAs which usually adopt the CE loss.

We note that the conventional hierarchy classifiers (e.g., HSRM) usually have very limited help for the accuracy of vanilla CE-loss, and also utilize a sophisticated structure and lead to a slower inference. Noticing that our DOT only uses the leaf level branch in the testing stage after the training, which has the same inference speed as the vanilla CE-loss version. 

$leaf$-$only$ and $equally$ Indicate the DOT training on leaf-level only and weighting each level equally respectively. We use the integrating scheme based on information gain unless specific notation. By inheriting the hierarchy in each level with a multi-task framework, the DOT can always gain a better TIE and accuracy than its $leaf$-$only$ or $equally$ DOT counterpart. Considering that the number of level in ImageNet is much more than the other datasets, the multi-task integration can boost the performance by a large margin.



In the large scale Imagenet 10K dataset, our DOT can still improve the performance. As shown in Table \ref{tab:3}, with the prior of semantic hierarchy, DOT can significantly outperforms CE loss and LMM which based on visual hierarchy.This task also indicates that our method can be a general alternative objective of CE loss and be applied in a plug and play fashion. 


\section{Conclusions}

We propose a novel risk minimization framework for image classification, namely discrete optimal transport (DOT) training, by leveraging semantic tree structure. We define the ground distance metric in the optimal transportation distance as the tree induced error and its increasing function. A hierarchical multi-task learning method is used to constrain the semantic correlation in different levels, and can be integrated with the information gain. Experimental results on several databases of semantic tree structures show that the proposed DOT method achieves superior performance.

\section{Acknowledgement}
This work was partially supported by the PolyU Central Research Grant (G-YBJW), Hong Kong Government General Research Fund GRF (Ref. No.152202/14E), and Jiangsu Youth Programme (BK20200238).

\bibliographystyle{IEEEbib}
\bibliography{IEEEabrv}

\end{document}